\documentclass[runningheads]{llncs}
\usepackage{graphicx}

\usepackage{tikz}
\usepackage{comment}
\usepackage{amsmath,amssymb} 
\usepackage{color}

\usepackage{booktabs}
\usepackage{multirow}
\usepackage[title]{appendix}
\usepackage{enumitem}
\usepackage{xspace}
\usepackage{balance}
\usepackage{mathtools}
\usepackage{rotating}
\usepackage{tabulary}
\usepackage{array}
\usepackage{subcaption}
\usepackage[linesnumbered,ruled]{algorithm2e}

\usepackage[accsupp]{axessibility}  

\newcommand{\lyx}[1]{\textcolor{black}{#1}}
\newcommand{\lyxt}[1]{\textcolor{black}{#1}}
\newcommand{\eccv}[1]{\textcolor{black}{#1}}
\newcommand{\mm}[1]{\textcolor{black}{#1}}
\newcommand{\lyxm}[1]{\textcolor{black}{#1}}

\begin{document}
\pagestyle{headings}
\mainmatter
\def\ECCVSubNumber{13}  

\title{Learning from Noisy Labels with Coarse-to-Fine Sample Credibility Modeling}


\titlerunning{CREMA: Coarse-to-fine learning with noisy labels}
%
\author{Boshen Zhang\inst{1}* \and
Yuxi Li\inst{1}* \and
Yuanpeng Tu\inst{2}* \and
Jinlong Peng\inst{1} \and
Yabiao Wang\inst{1}$^\dag$ \and
Cunlin Wu\inst{3} \and
Yang Xiao\inst{3} \and
Cairong Zhao\inst{2}
}
\authorrunning{Boshen Zhang et al.}
%
\institute{YouTu Lab, Tencent, Shanghai 
\email{\{boshenzhang,yukiyxli,jeromepeng,caseywang\}@tencent.com} \and
Tongji University, Shanghai \\
\email{\{2030809, zhaocairong\}@tongji.edu.cn}\\
 \and
Key Laboratory of Image Processing and Intelligent Control, Ministry of Education,\\School of Artificial Intelligence and Automation, \\
Huazhong University of Science and Technology, China\\
\email{\{cunlin\_wu,Yang\_Xiao\}@hust.edu.cn}}
\maketitle

\let\thefootnote\relax\footnotetext{
* Authors contributed equally to this work. \\
\dag Yabiao Wang is corresponding author (caseywang@tencent.com).}

\begin{abstract}
  Training deep neural network~(DNN) with noisy labels is practically challenging since inaccurate labels severely degrade the generalization ability of DNN.
 Previous efforts tend to handle part or full data in a unified denoising flow \mm{via identifying noisy data with a coarse small-loss criterion} to mitigate the \lyxm{interference from noisy labels, ignoring the fact that the difficulties of noisy samples are different, thus a rigid and unified data selection pipeline cannot tackle this problem well} . \mm{In this paper, we first propose a coarse-to-fine robust learning method called \textit{CREMA}, to handle noisy data in a divide-and-conquer manner. In coarse-level, clean and noisy sets are firstly separated \lyxm{in terms of credibility in a statistical sense}. 
 Since it is practically impossible to categorize all noisy samples correctly, we further process them in a fine-grained manner via modeling the credibility of each sample. Specifically, }
 for the clean set, we deliberately design a memory-based modulation scheme to dynamically adjust the contribution of each sample in terms of its historical credibility sequence during training, thus alleviating the effect from \lyxm{noisy samples incorrectly grouped into the clean set}. Meanwhile, for samples categorized into the noisy set, \mm{a selective label update strategy is proposed to correct noisy labels while mitigating the problem of correction error.}
 Extensive experiments \lyxm{are conducted on benchmarks of different modality, including image classification (CIFAR, Clothing1M etc) and text recognition (IMDB), with either synthetic or natural semantic noises, demonstrating the superiority and generality of \textit{CREMA}.}
\end{abstract}

\begin{figure*}[!t]
 \centering
\small
 \begin{minipage}{0.31\textwidth}
	\centering
	\includegraphics[width=\textwidth]{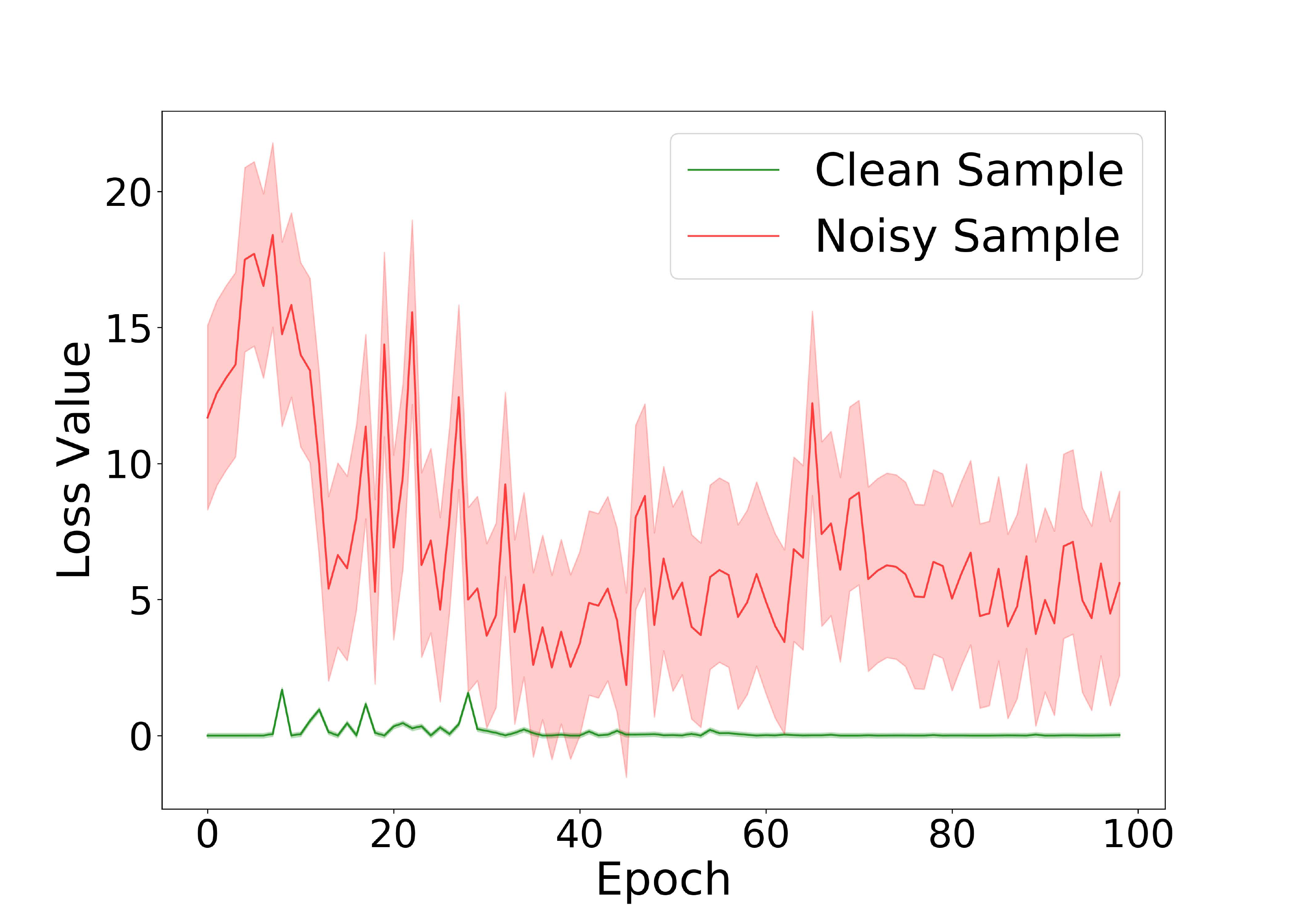}
	\vspace{-.4em}
	\scriptsize{(a)}	
\end{minipage}
\vspace{-.3em}
 \begin{minipage}{0.32\textwidth}
	\centering
	\includegraphics[width=\textwidth]{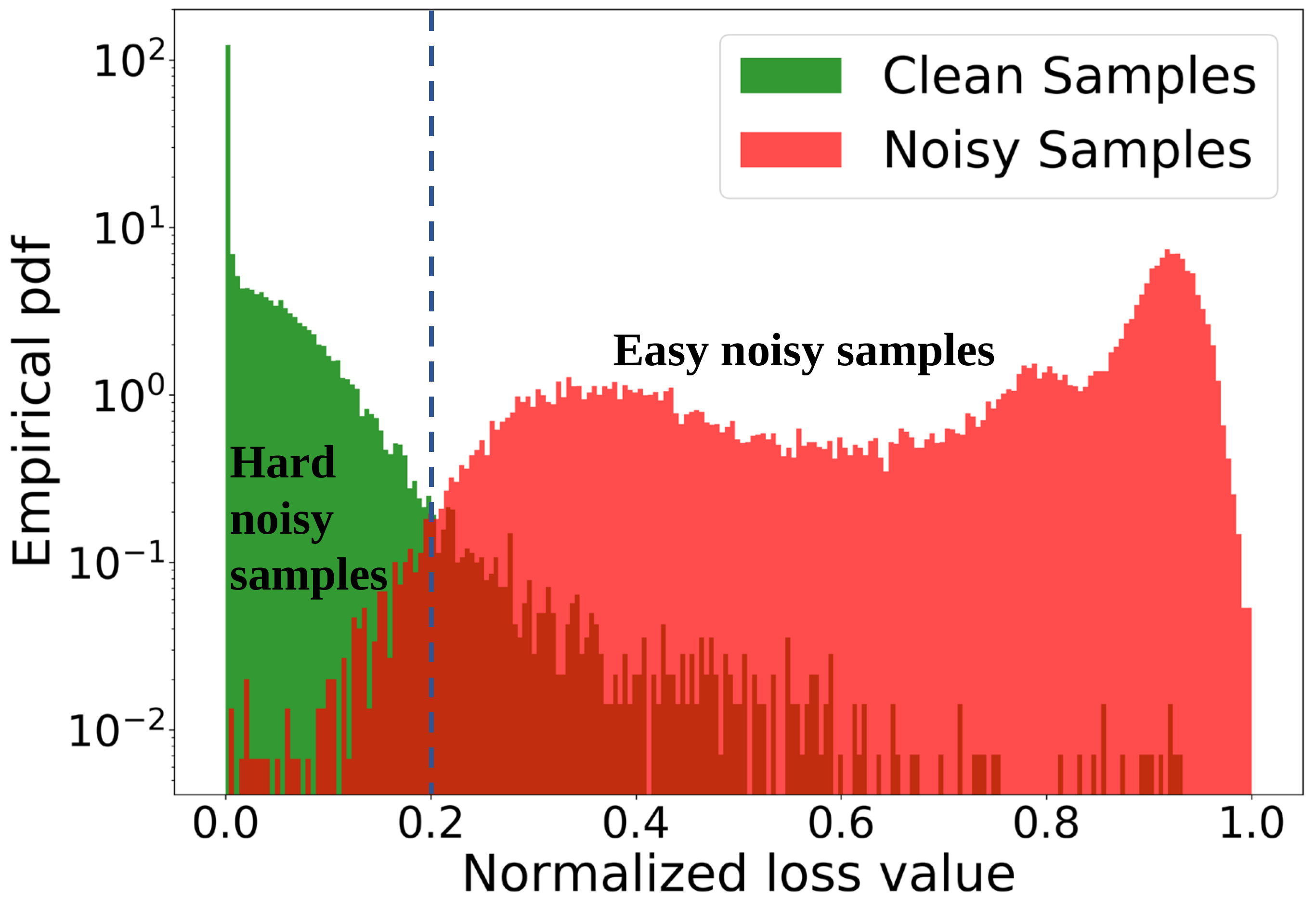}	
	\vspace{-.2em}
	\scriptsize{(b)}	
\end{minipage}
\vspace{-.3em}
 \begin{minipage}{0.32\textwidth}
	\centering
	\includegraphics[width=\textwidth]{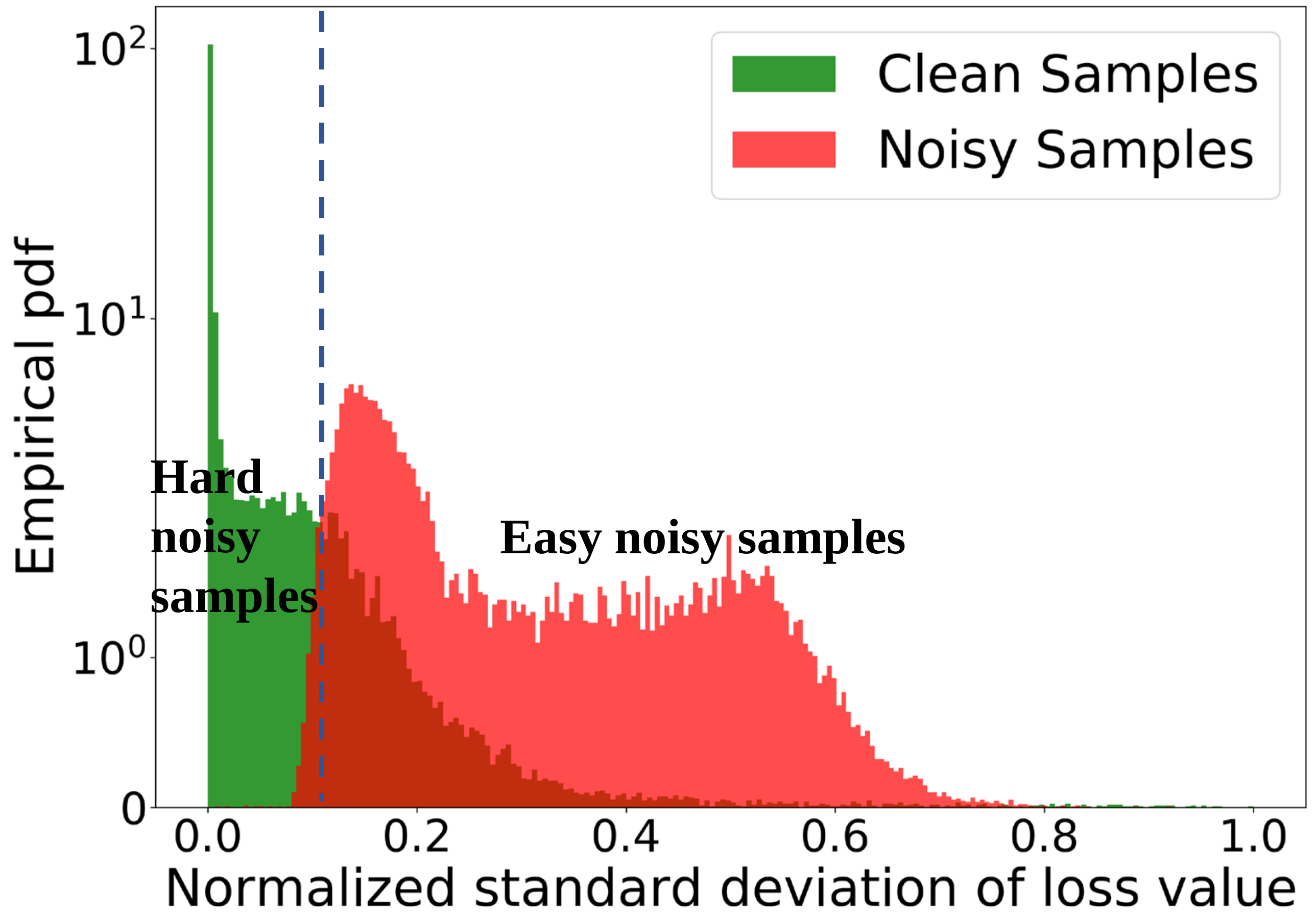}	\vspace{-.2em}
	 \scriptsize{(c)}	
\end{minipage}
\vspace{-.3em}
  \caption
  	{
  	\small
			Training on MNIST with 50\% symmetric noise. (a) Compared with noisy samples, clean samples yield relatively smaller loss value and more consistent predictions. (b) Empirical PDF~(Probability Density Function) of loss values and (c) their standard deviation justify the above conclusion. That is, clean and noisy samples possess distinctive statistical properties. However, noisy samples can not be completely identified via a simple threshold filter strategy (blue dotted line in (b) and (c)) with these statistical metrics. The existence of easy and hard noisy samples requires different ways to handle them accordingly. A similar experimental conclusion can also be found on other synthetic noisy datasets~(i.e., CIFAR-10, CIFAR-100) across different noise settings.
	  } 
  \label{fig:loss_magnitude}
\vspace{-1em}
 \end{figure*}
 
\section{Introduction}
Deep learning has achieved significant progress in \lyxm{the recognition of multimedia signals (e.g. images, text, speeches)}. The key to its success is the availability of large-scale datasets with reliable \lyxm{manual} annotations. Collecting such datasets, however, is time-consuming and expensive. \lyxm{Some alternative ways} to obtain labeled data, such as web crawling~\cite{Clothing1M}, inevitably yield samples with noisy labels, \lyx{which \mm{are} not appropriate to be directly utilized to} train DNN since these complex models can easily over-fitting~(i.e., memorizing) noisy labels~\cite{memorizationEffect,Zhang2017UnderstandingDL}.

To handle this problem, classical Learning with Noisy Label (LNL) approaches focus on either identifying \lyx{and dropping} noisy samples (i.e., sample selection)~\cite{Co-teaching,MentorNet,Co-teaching-PP,JOCOR} or \lyx{adjusting the objective term of} each sample during training (i.e., loss adjustment)~\cite{forward-correction,PENCIL,tanaka2018joint}. The former usually make use of small-loss trick to select clean samples, and then \lyx{take} them to update DNNs. However, the procedure of sample selection cannot guarantee that the selected clean samples are \lyx{completely} clean. \lyx{In contrast, as indicated in Fig.~\ref{fig:loss_magnitude}, division relied on statistic metrics can still involve some hard noisy samples in the training set, which will be treated equally as other normal samples in the following training stages.} 
Thus the negative impact brought by wrongly grouped noisy samples can still confuse the optimization process and lower the test performance of DNNs~\cite{Co-teaching-PP}. \lyx{On the other hand,} the latter \lyx{schemes} reweight loss values or update labels by estimating the confidence \lyxt{on how clean a sample is}. Typical methods include loss correction via an \lyx{estimated} noise transition matrix~\cite{forward-correction,noise-adaption-layer,gold-loss-correction}. However, estimating an accurate noise transition matrix is practically challenging.
\lyx{Recently, there are} approaches directly correcting the labels of all training samples~\cite{tanaka2018joint,PENCIL}. \lyx{However, we empirically find that unconstrained label correction in full data can do harm to clean samples and reversely hinder the model performance.}

Towards the problems above, we propose a simple but effective 
method called \textit{CREMA} (\textit{\mm{Coarse-to-fine} sample cREdibility Modeling and Adaptive loss reweighting}), \lyx{which} adaptively reduces the impact of noisy labels via modeling the credibility~(i.e., quality) of each sample. \mm{In the coarse-level,} with the estimated sample credibility \mm{by simple statistic metrics}, clean and noisy samples can be \mm{roughly} separated \lyx{and handled in a divide-and-conquer manner}. \lyx{Since} it is practically impossible to separate these samples perfectly, 
for the selected clean samples, we \lyx{take} their \lyx{historical credibility sequences} to adjust the contribution of each sample \lyx{to their objective}, thus mitigating the negative impact of hard noisy samples~\mm{(i.e, \lyxm{noisy} samples incorrectly grouped into the clean set) in a fine-grained manner}. As for the separated noisy samples, some of them are actually clean~(i.e., hard \mm{clean} samples) and can be \lyxm{helpful for model training}. Thus instead of discarding them as previous sample selection methods~\cite{Co-teaching,JOCOR}, we make use of them via a selective \lyx{label correction} scheme. 

\lyx{The insight behind CREMA is from the} observation \lyx{on the loss value during training on noisy data (illustrated in Fig~\ref{fig:loss_magnitude}), it can be found} that \textbf{\lyx{clean and noisy samples manifest distinctive statistical properties during training, where} clean samples \lyx{yield} relatively smaller loss value}~\cite{survey-bootstrapping} \textbf{and \lyx{more consistent prediction}}. Hence these statistical features can be utilized  
to \lyxm{coarsely} model the sample credibility. However, \lyx{Fig~\ref{fig:loss_magnitude} also shows that the full data can not be perfectly separated by simple statistical metrics. This inspires us to adaptively cope with noises of different difficulty levels \lyxm{with more fine-grained design}. For easily recognized noisy samples, we can directly apply certain label correction schemes while avoiding erroneous correction on normal samples. For} samples that fall into the confusing area \lyx{and hybrid with clean ones}, \lyx{since \lyxm{the coarsely} estimated credibility in the current epoch is not informative enough to identify noisy samples, \textit{CREMA} \lyxm{applies a fine-grained likelihood estimator of noisy samples by resorting the} historical sequence of sample credibility. } 
\lyx{This is achieved by maintaining a historical memory bank along with the training process and estimating the likelihood function through a consistency metric and assumption of markov property of the sequence.}

\textit{CREMA} is built upon a classic co-training framework~\cite{JOCOR,Co-teaching}. The \mm{fine-grained} sample credibility estimated by one network is used to adjust the loss term of credible samples for the other network. 
\lyxm{Extensive experiments are conducted on benchmarks of different modality, including image classification (CIFAR, MNIST, Clothing1M etc) and text recognition (IMDB), with either synthetic or natural semantic noises, demonstrating the superiority and generality of} the proposed method. In a nutshell, the key contributions of this paper include:

$\bullet$ \textit{CREMA}: a novel LNL algorithm that combats noisy labels via \mm{coarse-to-fine sample credibility modeling. In coarse-level, clean and noisy sets are roughly separated and handled respectively, in the spirit of the idea of divide-and-conquer. Easily recognized noisy samples are handled via a selective label update strategy;}

$\bullet$ \lyx{In \textit{CREMA}, likelihood estimation of historical credibility sequence is proposed to help identify hard noisy samples, which naturally plays as the dynamical weight to modulate loss term of each training sample~\mm{in a fine-grained manner};}

$\bullet$ \textit{CREMA} is \lyxm{evaluated} on \lyxm{six} synthetic and real-world noisy datasets \lyxm{with different modality, noise type and strength}. Extensive ablation studies and qualitative analysis are provided to verify the effectiveness of each component. 



\begin{figure*}[t]
\centering
 \includegraphics[width=0.98\textwidth]{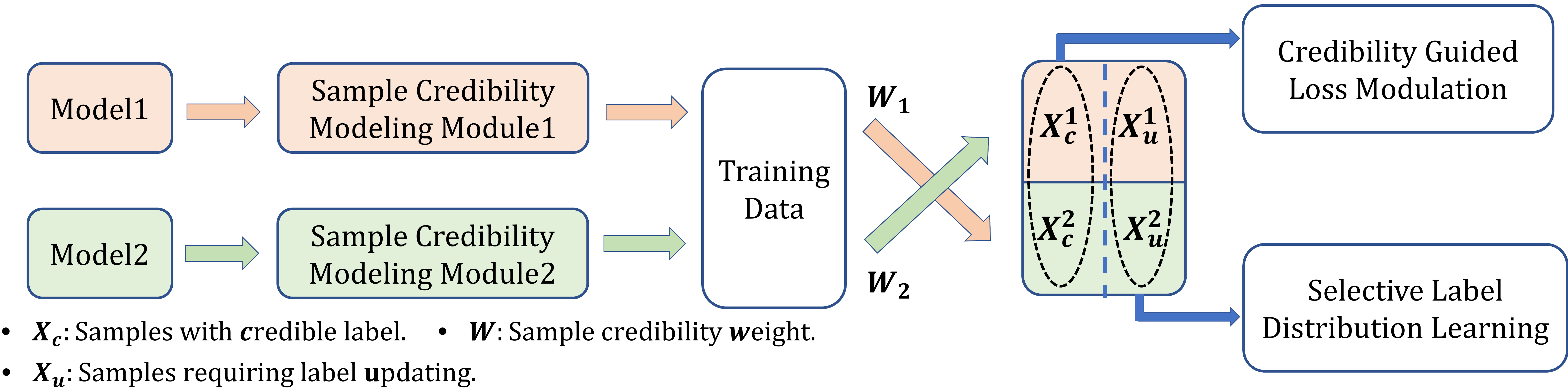}
 \vspace{-.5em}
  \caption{
  \small
   The pipeline of \textit{CREMA}. \textit{CREMA} trains two parallel networks simultaneously. Clean samples (mostly clean) $X_c$ and noisy samples (mostly noisy) $X_u$ are separated via estimating the credibility of each training data. A selective label distribution learning scheme is applied for easily distinguishable noisy samples in $X_u$. As for the clean set $X_c$, likelihood estimation of historical credibility sequence is proposed to handle the hard noisy samples via adaptively modulating their loss term during training. 
  }
\label{fig:Pipeline}
\end{figure*}

\section{Related Works} \label{Related Works}
The existing LNL approaches can be mainly categorized into three groups: loss adjustment, label correction, and noisy sample detection. Next, we will introduce and discuss existing works for training DNN with noisy labels.

\textbf{Loss Adjustment.} 
Adjusting the loss values of all training samples is able to reduce the negative impact of noisy labels. To do this, many approaches seek to robust loss functions, such as Robust MAE~\cite{Robust-MAE}, generalized cross entropy~\cite{GCE}, symmetric cross entropy~\cite{SCE}, Improved MAE~\cite{IMAE} and curriculum loss~\cite{Curriculum-Loss}. Rather than treat all samples equally, some methods rectify the loss of each sample through estimating the label transition matrix~\cite{gold-loss-correction,forward-correction,noise-adaption-layer,masking,Clothing1M,transition_matrix_wo_clean_data_1,transition_matrix_wo_clean_data_2} or imposing different importance on each sample to formulate a weighted training procedure~\cite{importance-reweight-1,importance-reweight-2,active-bias,DM}. The noise transition matrix, however, is relatively hard to be estimated and many approaches~\cite{gold-loss-correction,C2D-minimal,C2D-Soseleto,MentorNet,survey-Knowledge-Distlling,survey-L2LWS,survey-CWS,survey-automatic-reweighting,survey-meta-weight-net,CVPRzhang2020distilling} often make assumptions that a small clean-labeled dataset exists. In real-world scenarios, such condition is not always fulfilled, thus limiting the applications of these approaches.

\textbf{Label correction.}
Label correction methods seek to refurbish the ground-truth of noisy samples, thus preventing DNN overfits to false labels. The most common ways to obtain the updated label include bootstrapping~(i.e., a convex combination of the noisy label and the DNN prediction)~\cite{survey-bootstrapping,iccv19-deep-self,Arazo-BMM,wang2021proselflc} and label replacing~\cite{tanaka2018joint,PENCIL,SELFIE,PLC}. One critical problem of 
label correction methods is to define the confidence of each label being clean, that is, samples with high clean probability should keep their labels almost unchanged, and vice versa. Previous solutions including cross-validation~\cite{survey-bootstrapping}, fitting a two-component mixture model~\cite{Arazo-BMM}, local intrinsic dimensionality measurement~\cite{Local-intrinsic-dimensionality,survey-D2L} and leveraging the prediction consistency of DNN models~\cite{SELFIE,wang2021proselflc}. However, updating the labels of all training sets is challenging, and well-designed regularization terms are important to prevent DNN from falling into trivial solutions~\cite{tanaka2018joint,PENCIL}.

\textbf{Noisy sample detection.}
One common knowledge used to discover noisy samples is the \textit{memorization effects} ~(i.e., DNN fits clean samples first and then noisy ones). As a result, after a warm-up training stage with all noisy samples, DNN is able to identify the clean samples by taking the small-loss ones. The~\textit{small-loss trick} is exploited by many sample selection methods~\cite{Co-teaching,Co-teaching-PP,Decouple,MentorNet,JOCOR,SELF,SELFIE,S2E}. After separating the noisy samples from the clean ones, Co-teaching~\cite{Co-teaching} and \lyxt{corresponding} variants~\cite{Co-teaching,Co-teaching-PP,JOCOR,S2E} update two parallel network parameters with the clean samples and abandoned the noisy ones. The idea of training two deep networks simultaneously is effective to avoid the confirmation bias problem~(i.e., a model would accumulate its error through the self-training process)~\cite{Co-teaching,MentorNet,dividemix}. 
Other \mm{noisy measurement} metrics such as Area Under the Margin (AUM)~\cite{AUM-NIPS2020} is also proposed to better distinguish and remove noisy samples, which hypothesizes that noisy samples will in expectation have smaller AUM \lyxt{than} clean samples.
\lyxt{However}, discarding the noisy samples means that valuable data may be lost, which leads to slow convergence of DNN models~\cite{active-bias}. Instead, there are methods that utilize both clean and noisy samples to formulate a semi-supervised learning problem, by discarding only the labels of identified noisy samples. Thus \mm{naturally} converting LNL problem into a semi-supervised learning one, for which powerful semi-supervised learning methods can be leveraged to boost performance~\cite{dividemix,C2D,mixup,mixmatch}.

\textbf{Hybrid.}
There are \lyxt{also researches} taking two or more techniques \lyxt{above} into account to boost performance \lyxt{of robust learning}. \lyxt{For example,} RoCL~\cite{RoCL_iclr2021} and SELFIE~\cite{SELFIE} propose to dynamically discover informative~(refurbished) samples and correct their labels with model predictions.
Accordingly, \textit{CREMA} belongs to hybrid group and it differs from existing methods in (1) it adaptively cope with noises of different difficulty levels via \mm{estimating the sample credibility in a coarse-to-fine fashion, easy and hard noisy samples are handled in a divide-and-conquer} strategy; (2) it estimates likelihood of historical sample credibility sequence to dynamically modulate loss term of hard noisy samples; (3) it explores a selective label correction scheme to deal with hard clean samples \mm{while mitigating the correction error}; 
(4) it is end-to-end \lyxt{trainable} and does not require extra computation or any modification to the model.


\begin{figure*}[!t]
 \centering
\small
 \begin{minipage}{0.49\textwidth}
	\centering
	\includegraphics[width=\textwidth]{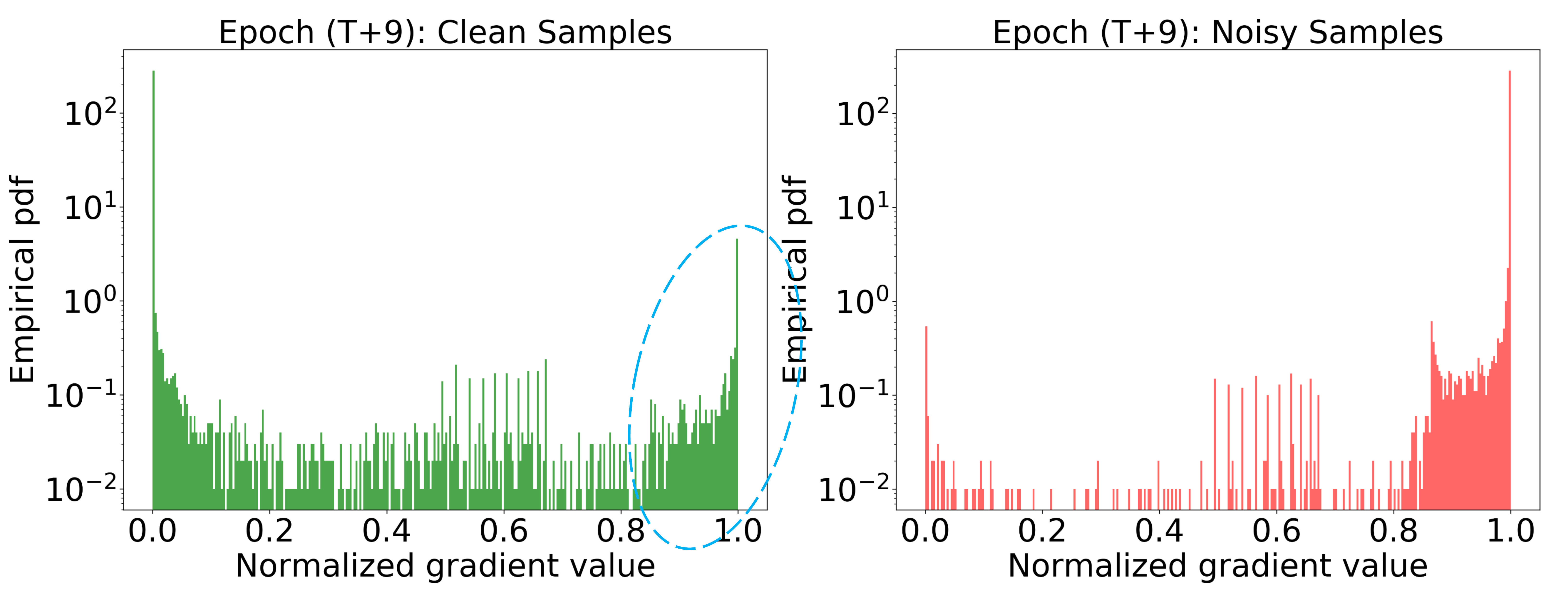}
	(a)	{\small Global label learning.}
\end{minipage}
 \begin{minipage}{0.49\textwidth}
	\centering
	\includegraphics[width=\textwidth]{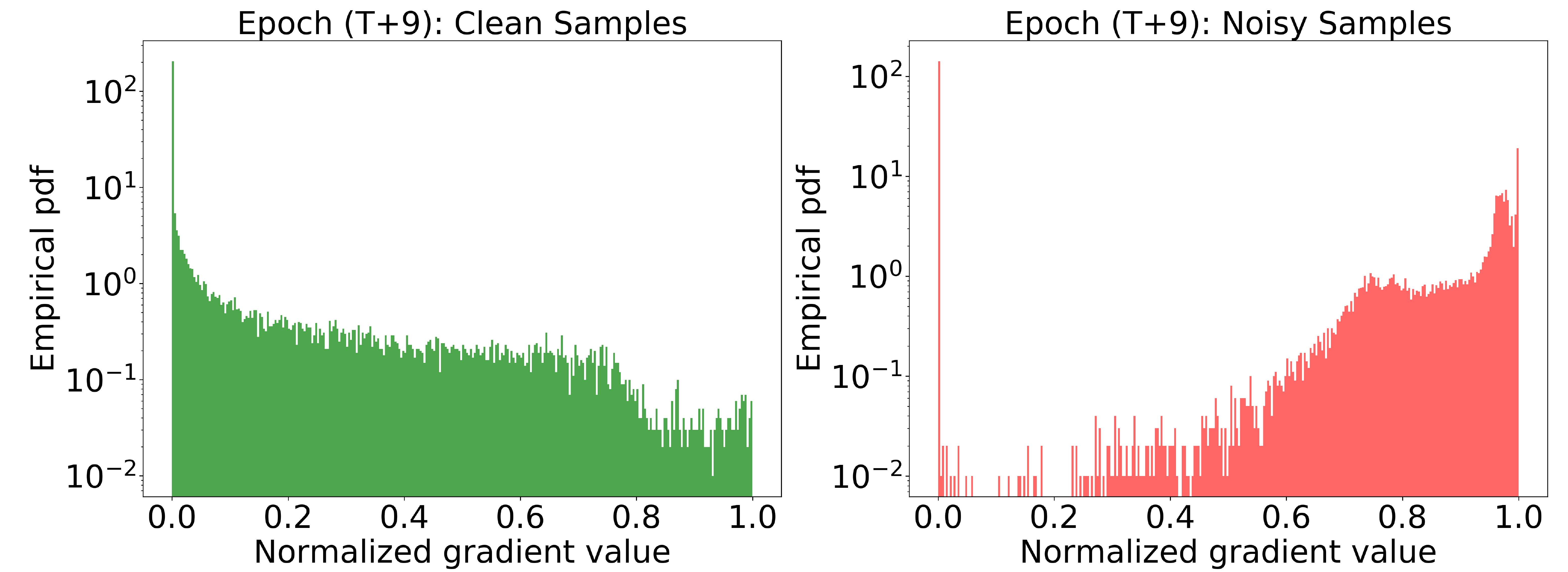}	
	(b)	{\small Selective label learning in \textit{CREMA}.}
\end{minipage}
\vspace{-1em}
  \caption
  	{
  	\small
		Training on MNIST with 50\% symmetric noise, warm up (i.e., training on all samples with original noisy labels) for T epochs. (a) Global learning procedure requires updating all training samples' label, causes relatively large gradient values even on clean samples~(areas within blue dotted lines), making it hard to focus on correcting noisy labels. (b) Training on \textit{CREMA} can effectively identify noisy samples and focus on correcting noisy labels with relatively large gradient values.
	  } 
\vspace{-1em}
\label{fig:grad}

\end{figure*}

\section{Method} \label{Method}
In this section, we introduce \textit{CREMA}, an end-to-end approach for LNL problem. The technical pipeline of the approach is shown in Fig.~\ref{fig:Pipeline}. \lyx{The training process is built upon a classic} co-training framework~\cite{Co-teaching,JOCOR} to avoid confirmation bias and separate credible samples (mostly clean) and noisy samples (mostly noisy) via per-sample loss value \mm{in the coarse stage}. \mm{The separated samples are handled by the succeeding fine-grained processes in a divide-and-conquer manner.}

\subsection{Coarse level separation}   \label{Method:coarse}

Formally, for multi-class classification with noisy label problem, let {\small $\mathcal{D}=\{(x_i,y_i)\}_{i=1}^N$} denote the training data, where $x_i$ is \lyxm{data sample} and {\small $y_i \in \{0,1\}^C$} is the one-hot label over $C$ classes. $f(x_i;\theta)$ denotes \lyxm{sample} feature extracted by DNN model. With the loss $\mathcal{L}(x, y)$ from the DNN model, clean set $\mathcal{X}_c$, and noisy set $\mathcal{X}_u$ are separated via the widely used low-loss criterion~\cite{Co-teaching,JOCOR},
\begin{equation}
\begin{aligned}\label{eq:coarse_sep}
& \mathcal{X}_c = \{(x, y) | \mathcal{L}(x, y) < \tau , (x, y) \in \mathcal{D}\}, \\
& \mathcal{X}_u = \{(x, y) | \mathcal{L}(x, y) \ge \tau , (x, y) \in \mathcal{D}\},  
\end{aligned}
\end{equation} 
\lyx{where $\tau$ is the threshold, determined by a} dynamic \lyx{memory rate} $R(t) \in [0, 1]$. \lyx{Which is set for DNN to gradually distinguish $(1-R(t))$ data with highest loss value} as noisy samples \lyx{while keeping other samples as the clean set}. \mm{The loss value simply serves as credibility of each sample in a coarse level. However, as illustrated in Fig. \ref{fig:loss_magnitude}, this simple separation criterion can not \lyx{strictly eliminate noisy samples}.}
Hence we choose to handle them respectively, where  $\mathcal{X}_c$ \lyxt{is exploited to} update DNN parameters via \mm{fine-grained} sample credibility guided loss adjustment~(Sec. \ref{Method:sub1}), and $\mathcal{X}_u$ is leveraged via another label learning \lyx{scheme}~(Sec. \ref{Method:sub2}).

\subsection{Fine-grained sequential credibility modeling}   \label{Method:sub1}


\lyx{\textbf{Sequential credibility analysis}. 
Previous works prefer to assess the data reliability purely based on its statistical property on a single point of time (e.g. the loss value in current epoch) during training process, i.e. they regard the credibility $w(x, y)$ of the $i$-th data sample $(x, y)$ proportional to the joint distribution \lyxt{or likelihood} of its data-label pair,}
\begin{equation}
    w(x, y) \propto P(x, y) \quad \textbf{or} \quad  w(x, y) \propto \log P(x, y).
\end{equation}
However, as shown in Fig.~\ref{fig:loss_magnitude}, \lyx{the training curve of normal and noisy samples usually yield different statistic information, where} noisy samples usually have relatively larger loss values and poorer prediction consistency compared with clean ones, \lyx{therefore the historical record of data training is also informative enough to help distinguish noisy and clean data.} 

\lyx{This observation inspires us to estimate the data credibility in a sequential manner. To be specific, we define a sequence with length $n$ as:
\begin{equation}\label{eq:sequence}
    \mathbf{L}^n_t=[\mathbf{f}_t,\mathbf{f}_{t-1},\cdots,\mathbf{f}_{t-n+1}], \quad \mathbf{f}_t = f\left(x; \theta_t\right).
\end{equation}
Eq~(\ref{eq:sequence}) illustrates a sliding window covering the feature snapshot of data from previous $n$ epochs to current time point, where $\theta_t$ denotes the model parameters at the $t$-th epoch, and we model the data credibility with the likelihood and consistency of these historical sequences,
\begin{equation}\label{eq:credibility}
    w(x, y) \propto C(\mathbf{L}^n_t, y)\log{P\left(\mathbf{L}^n_t|\mathbf{f}_{t-n}, y\right)}.
\end{equation}
The Eq~(\ref{eq:credibility}) can be decoupled \lyxm{into} two items, where $C(\mathbf{L}^n_t, y)$ measures the stability of training sequence given its label $y$, while $\log{P\left(\mathbf{L}^n_t|\mathbf{f}_{t-n}, y\right)}$ denotes log-likelihood of sequence generated from the $(t-n)$-th data observation of neural network training process. To estimate the sequential log-likelihood, we further assume that the observation in sequence $\mathbf{L}^n_t$ conforms to a certain markov property as:
\begin{equation}\label{eq:markov}
    \mathbf{f}_t \perp \mathbf{f}_i | (\mathbf{f}_{t-1}, y) \quad \forall\quad i < t-1.
\end{equation}
The assumption in Eq~(\ref{eq:markov}) is reasonable since in most iterative learning algorithm like SGD, the data feature distribution is only decided by the last observation and its label. With this assumption, we can further derive the likelihood as:
\begin{equation}
\begin{aligned}\label{eq:derive}
\log{P\left(\mathbf{L}^n_t|\mathbf{f}_{t-n}, y\right)} & = \log{P\left(\mathbf{f}_t|\mathbf{L}^{n}_{t-1}, y\right)} + \log{P\left(\mathbf{L}^{n-1}_{t-1}|\mathbf{f}_{t-n}, y\right)} \\
& = \sum_{i=0}^{n-1}{\log{P\left(\mathbf{f}_{t-i}|\mathbf{L}^{n-i}_{t-i-1}, y\right)}} \\
& = \sum_{i=0}^{n-1}{\log{P\left(\mathbf{f}_{t-i}|\mathbf{f}_{t-i-1}, y\right)}}.
\end{aligned}
\end{equation}
With Eq~(\ref{eq:derive}), we can represent the sequential likelihood as the summation of the conditional likelihood of data observation at each \lyxt{adjacent epochs} within a sliding window of length $n$. In the implementation, we can apply a normalized mixture model like GMM~\cite{dividemix} or BMM~\cite{Arazo-BMM} as estimator to estimate the conditional probability $P\left(\mathbf{f}_{t-i}|\mathbf{f}_{t-i-1}, y\right)$ in Eq~(\ref{eq:derive}) via modeling the sample-wise loss value distribution. Meanwhile, with the conditional probability estimation, the stability measurement $C(\mathbf{L}^n_t, y)$ is further designed as a modulator to suppress loss on training sequence with intense fluctuation,
\begin{align}\label{eq:variance}
    C(\mathbf{L}^n_t, y) & = 1 -  \sqrt{\frac{1}{n}\sum_{i=0}^{n-1}{\left(P\left(\mathbf{f}_{t-i}|\mathbf{f}_{t-i-1}, y\right) - \Bar{P}\left(\mathbf{L}^n_t, y\right)\right)^2}}.
\end{align}
}
\begin{align}\label{eq:mean}
    \Bar{P}\left(\mathbf{L}^n_t, y\right) & = \frac{1}{n}\sum_{i=0}^{n-1}P\left(\mathbf{f}_{t-i}|\mathbf{f}_{t-i-1}, y\right).
\end{align}

\textbf{Adaptively loss adjustment.} 
The sequential likelihood $\Bar{P}\left(\mathbf{L}^n_t\right)$ and stability measurement $C(\mathbf{L}^n_t, y)$ reflects how confident of the sample being clean. With the estimated credibility we reweight loss to update DNN as:

\begin{equation}
\label{eq:loss_reweighting}
{\theta}_{t+1} ={\theta}_{t} - \eta\nabla \Big( \frac{1}{|\mathcal{X}_{c}|}\!\sum_{(x,{y})\in\mathcal{X}_c}\!\!\!\!{w(x,{y}) \mathcal{L} \big(f(x;\theta_t), {y}\big)}\Big).
\end{equation} 

Where $\mathcal{L}$ is the objective function. $w(x,{y})$ is the sample credibility and it modulates the contribution of each sample through gradient descending algorithm. Note that Eq~(\ref{eq:loss_reweighting}) is only applied on clean set $\mathcal{X}_c$, in this way, the negative impact of hard noisy samples within $\mathcal{X}_c$ can be mitigated.

\textbf{Objective function.} 
Inspired by the design of symmetric cross entropy (SCE) function~\cite{SCE}, a symmetric JS-divergence function with a co-regularization term is leveraged in \textit{CREMA} as:

\begin{equation}
\begin{aligned}\label{eq:loss}
\mathcal{L} & = D_{\mathrm{JS}}(y || h(f_1(x; \theta))) + D_{\mathrm{JS}}(y || h(f_2(x; \theta))) \\
& + D_{\mathrm{JS}}\left( h(f_1(x; \theta)) || h(f_2(x; \theta))\right),
\end{aligned}
\end{equation} 

Where $h(x)$ is the softmax function, $f_1(x)$ and $f_2(x)$ are features extracted by two models. The reason we choose JS-divergence~\cite{foundations1999} instead of cross entropy~(CE) as loss function is that CE tends to over-fit noisy samples as these samples contribute relatively large gradient values during training. While JS-divergence mitigates this problem via using predictions of the current model as supervising signals as well. Since for noisy samples, DNN predictions are usually more reliable than its label. Following previous work~\cite{tanaka2018joint}, a prior label distribution term and a negative entropy term are included to regularize training and further alleviate the over-fitting problem.

\begin{algorithm}[!t]
	\DontPrintSemicolon
	\small
	\textbf{Input:} network parameters $\theta^{(1)}$ and $\theta^{(2)}$, training dataset $\mathcal{D}$,  dynamic memory rate $R(t)$, soft label distribution $\tilde{{y}}$, memory sequence $\mathbf{L}^n_{1,t}$ and $\mathbf{L}^n_{2, t}$. \leavevmode \\
	\While{$t<\mathrm{MaxEpoch}$}    
	{	
	Fetch mini-batch $\mathcal{D}_n$ from $\mathcal{D}$; \leavevmode \\
	Divide $\mathcal{D}_n$ into $\mathcal{X}_c$ and $\mathcal{X}_u$ based on $R(t)$; \tcp*{divide samples into clean and noisy set based on low-loss criterion}
    \For {$x_c \in \mathcal{X}_c$}
    {
        Calculate $w(x_c,y_c)$ based on Eq~(\ref{eq:derive}) and Eq~(\ref{eq:variance}); \tcp*{sample credibility modeling} \leavevmode \\
        Update $\theta^{(1)}$ and $\theta^{(2)}$ based on Eq.~(\ref{eq:loss_reweighting}); \tcp*{adaptive loss adjustment}
	}
	\For {$x_u \in \mathcal{X}_u$}
	{
	    Update $\tilde{{y}}$, $\theta^{(1)}$, and $\theta^{(2)}$ through gradient descent;  \tcp*{update soft label distribution and model parameters}
	}   
    Update ${R(t)}$; \leavevmode \\
	Update $\mathbf{L}^n_{1,t}$ and $\mathbf{L}^n_{2, t}$; \tcp*{enqueue feature snapshot of current epoch}
	}
	\textbf{Output:} $\theta^{(1)}$ and $\theta^{(2)}$.
	\caption{\small \textit{CREMA}. Line 5-9: sequential credibility modeling; Line 10-12: selective label update.}
	
	\label{alg:crema}
\end{algorithm}
\vspace{-5mm}

\subsection{Selective label distribution learning}   \label{Method:sub2}

Following the divide-and-conquer idea, we attempt to leverage the separated noisy samples $\mathcal{X}_u$ as well. Some hard clean samples are blended with the separated noisy ones. Thus instead of discarding these wrongly labeled data as in most sample detection methods~\cite{Co-teaching,JOCOR}, we resort to label correction approaches~\cite{tanaka2018joint,PENCIL} to \lyx{exploit them with gradually corrected labels} and further boost performance.

Specifically, labels of $\mathcal{X}_u$ are treated as extra parameters and updated through back-propagation to optimize a certain objective, this means both the network parameters and labels are updated simultaneously during the training process, where original one-hot labels ${{y}}$ will turn into a soft label distribution $\tilde{{y}}=h(y)$ after updating. Formally, $\tilde{{y}}$ is updated as $\tilde{{y}} \leftarrow \tilde{{y}} - \lambda (\partial \mathcal{L}_{l} / \partial \tilde{{y}})$. Where $\lambda$ is learning rate and $\mathcal{L}_{l}$ is the objective to supervise the label correction process as $\mathcal{L}_{l} = D_{\mathrm{JS}}(h(f_1(x; \theta) || \tilde{{y}}) + D_{\mathrm{JS}}(h(f_2(x; \theta) || \tilde{{y}})$.

\textbf{Empirical insight.} In our experiments. we find that global label learning strategy (i.e., correcting labels of all training data) suffers from correction error in clean data. This can be observed from Fig.~\ref{fig:grad} (a), large gradient value will also be imposed on lots of correctly-labeled samples. Consequently, labels for these clean samples are unnecessarily updated. Compared with global label correction manners, we choose to only update the separated noisy samples $\mathcal{X}_u$ (mostly noisy). As shown in Fig.~\ref{fig:grad} (b), the proposed selective label correction strategy focuses more on learning noisy labels. The number of correctly-labeled samples with large gradient value is way less than a global correction scheme. 
\eccv{Indicating that the selective label correction scheme can mitigate the problem of correction error.}
Experiments in Sec~\ref{ablation} also quantitatively verify the effectiveness of the selective label learning strategy over the global label learning manner.

Putting this all together, Algorithm~\ref{alg:crema} delineates the proposed \textit{CREMA} in detail. In a nutshell, \textit{CREMA} is built on a divide-and-conquer framework. Firstly, clean set $\mathcal{X}_c$ and noisy set $\mathcal{X}_u$ are separated based on the low-loss criterion~\cite{Co-teaching}. For $\mathcal{X}_c$, we compute the likelihood of historical credibility sequence, which helps to adaptively modulate the loss term of each training sample. As for $\mathcal{X}_u$, a selective label correction scheme is leveraged to update label distribution and model parameters simultaneously. After each training epoch, memory sequence $\mathbf{L}^n_{1,t}$ and $\mathbf{L}^n_{2, t}$ are updated with the feature snapshot of the most current epoch.

\section{Experiments} \label{Experiments}
\subsection{Datasets and Implementation Details}
\textbf{Datasets.} To validate the effectiveness of the proposed method, we experimentally investigate on \mm{four} synthetic noisy datasets, i.e., \mm{IMDB~\cite{IMDB},} MNIST, CIFAR-10, CIFAR-100 \cite{CIFAR10} and two real-world label noise datasets, i.e., Clothing1M \cite{Clothing1M}, and Animal10N \cite{SELFIE}. \mm{IMDB~\cite{IMDB} is a collection of highly polarized movie reviews (positive/negative). It consists of 25,000 training samples and 25,000 samples for testing. The task is formalized as a binary classification task to decide the polarity of sentiment for a review.} MNIST consists of 70,000 images of size $28\times28$ for 10 classes, in which 60,000 images for training and the left 10,000 images for testing. Both CIFAR-10 and CIFAR100 contain 50,000 training images and 10,000 testing images of size $32\times32\times3$. Differently, the former has 10 classes, while CIFAR-100 has 100 classes. For the Clothing1M, it is a large-scale real-world noisy dataset which is collected from multiple online shopping websites. It contains 1 million training images and clean training subsets (47K for training, 14K for validation and 10K for test) with 14 classes. The noise rate for this dataset is around 38.5\%.
Animal-10N contains 55,000 human-labeled online images for 10 confusing animals. It includes approximately 8\% noisy-labeled samples. Following the settings in previous works~\cite{SELFIE,PLC}, 50,000 images are exploited as a training set while the left for testing. 

\textbf{Implementation Details.} For the three synthetic \mm{image classification} noisy datasets, \mm{MNIST, CIFAR-10 and CIFAR-100}, we follow the setting in previous works~\cite{Co-teaching,Co-teaching-PP,JOCOR}, experiments with three kinds of noise types are considered, i.e., symmetric noise (uniformly random), asymmetric noise, and pairflip noise. \mm{For the IMDB text classification dataset, we tokenize each sentence and the word embeddings have dimension 10,000. Symmetric noises with different noise level are tested.}
Specifically, symmetric noise is generated by replacing labels in each class with labels of other classes uniformly. Asymmetric noise simulates fine-grained classification (for example, lynx and cat in Animal-10N~\cite{SELFIE} with noisy labels, where labels are corrupted to a set of similar classes. Pairflip noise is generated by flipping each class to its adjacent class. Varying noise rates $\tau$ are conducted to fully evaluate the proposed method, where \mm{for symmetric label noise, we set} $\tau \in \{20\%, 50\%, 80\%\}$ \mm{on image datasets, $\tau \in \{20\%, 40\% \}$ on text dataset},
$\tau = 40\%$ for asymmetric noise and $\tau \in \{40\%, 45\%\}$ for pairflip label noise. For real-world noisy Clothing1M dataset, following ~\cite{PENCIL,PLC}, we do not use the 50K clean data, and a randomly sampled pseudo-balanced subset includes about 260K images is leveraged as training data.

For the network structure, \mm{a 2-layer bi-directional LSTM network is adopted for IMDB. It is of 128 embedding size and 128 hidden size.}
A 9-layer CNN with Leaky-ReLU activation function~\cite{Co-teaching} is used for MNIST, CIFAR-10, and CIFAR-100, while ResNet-50 is adopted for Clothing1M and Animal-10N datasets. The batch size is set as 64 for all the datasets. For fair comparisons, we train our model for 200 epochs in total and choose the average test accuracy of last 10 epochs as the final result in three \mm{image} synthetic noisy datasets. \mm{For IMDB dataset, we set total training epochs as 100 and also test the accuracy of last ten epochs.}
Total training epochs for Clothing1M and Animal-10N are 80 and 150 respectively. Additionally, all the methods are implemented in PyTorch and run on NVIDIA Tesla V100 GPUs. Moreover, we use Adam optimizer for all the experiments and set the initial learning rate as 0.001, then it is degraded by a factor of $5$ every $30$ epochs for Clothing1M and $50$ epochs for Animal-10N. The two classifiers in our methods are two networks with the same structure but different initialization parameters. Following~\cite{Co-teaching}, $R(t)$ is linearly decreased along with training until reach a lower bound value $\sigma$, for Clothing1M and Animal-10N datasets, we empirically set lower bound $\sigma$ as 0.8 and 0.92 respectively.

\begin{table*}[!tp]
\small
\centering
\scalebox{0.8}
{
\begin{tabular}{c | c | c  | c | c | c | c  }
\toprule
Noise rates $\tau $ & Standard & PENCIL  & Co-teaching & Co-teaching+ & JoCoR &CREMA (ours)\\
\midrule
Symmetry-20\% &  $79.94 \pm 0.10$ &  $97.20 \pm 0.53$   & $97.40 \pm 0.09$ & $97.81 \pm 0.03$ & $97.98 \pm 0.02$ & $\textbf{98.40} \pm 0.14$\\
\midrule
Symmetry-50\% &  $52.92 \pm 0.21$ &  $96.22 \pm 0.13$  & $92.47 \pm 0.14$ & $95.80 \pm 0.09$ & $96.35 \pm 0.02$ & $\textbf{98.07} \pm 0.24$\\
\midrule
Symmetry-80\% &  $23.95 \pm 0.18$ &  $87.64 \pm 0.25$    & $82.04 \pm 0.43$ & $58.92 \pm 0.37$ & $85.51 \pm 0.08$ & $\textbf{92.02} \pm 0.54$\\
\midrule
Asymmetry-40\% & $78.80 \pm 0.09$ &  $94.39 \pm 0.37$   & $90.57 \pm 0.04$ & $93.28 \pm 0.43$ & $94.14 \pm 0.12$ & $\textbf{97.15} \pm 0.26$ \\
\midrule
Pairflip-40\% &  $ 58.51\pm 0.29$ &  $94.06\pm 0.09$  & $90.73 \pm 0.22$ & $89.91 \pm 0.31$ & $93.47 \pm 0.10$& $\textbf{95.80} \pm 0.51$\\
\midrule
Pairflip-45\% &  $ 54.54\pm 0.30$ & $90.73 \pm 0.29$   & $89.42 \pm 0.22$ & $85.81 \pm 0.30$  & $91.30 \pm 0.25$ & $\textbf{94.12} \pm 0.58$\\

\bottomrule
\end{tabular}
}
\caption{\small Average test accuracy (\%) on \textit{MNIST} over the last ten epochs.}
\vspace{-1em}
\label{tab:mnist}
\end{table*}

\begin{table*}[!t]
\small
\centering
\scalebox{0.8}
{
\begin{tabular}{c | c | c  | c | c | c | c }
\toprule
Noise rates $\tau $ & Standard & PENCIL   & Co-teaching & Co-teaching+ & JoCoR & CREMA (ours)\\
\midrule
Symmetry-20\% &  $68.67 \pm 0.11$ & $78.78 \pm 0.15$  & $82.56 \pm 0.24$ & $82.27 \pm 0.21$ & $85.73 \pm 0.19$ & $\textbf{86.32} \pm 0.16$\\
\midrule
Symmetry-50\% &  $42.31 \pm 0.18$ &  $64.71 \pm 0.27$   & $72.97 \pm 0.22$  & $63.01 \pm 0.33$ & $79.53 \pm 0.10$ & $\textbf{81.63} \pm 0.13$\\
\midrule
Symmetry-80\% &  $15.94 \pm 0.07$ &  ${26.96} \pm 0.37$  & $24.03 \pm 0.18$ & $17.96 \pm 0.06$ & $27.30 \pm 0.08$ & $\textbf{29.66} \pm 0.16$ \\
\midrule
Asymmetric-40\% &  $70.04 \pm 0.08$ & $70.06 \pm 0.28$  & $75.96 \pm 0.15$  & $72.21 \pm 0.43$ & $76.31 \pm 0.21$ & $\textbf{82.49} \pm 0.13$ \\
\midrule
Pairflip-40\% &  $51.66 \pm 0.11$ &  $75.26 \pm 0.18$ &  $75.10 \pm 0.23$& $57.59 \pm 0.45$  & $68.56 \pm 0.16$ & $\textbf{85.00} \pm 0.13$\\
\midrule
Pairflip-45\% &  $45.78 \pm 0.13$ &  $71.18 \pm 0.28$   & $70.68 \pm 0.23$&  $49.60 \pm 0.23$ & $57.68 \pm 0.21$& $\textbf{82.94} \pm 0.12$\\
\bottomrule
\end{tabular}
}
\caption{\small Average test accuracy (\%) on \textit{CIFAR-10} over the last ten epochs.}
\vspace{-1em}
\label{tab:cifar10}
\end{table*}

\begin{table*}[!tp]
\small
\centering
\scalebox{0.8}
{
\begin{tabular}{c | c | c  | c | c | c | c}
\toprule
Noise rates $\tau $ & Standard & PENCIL & Co-teaching & Co-teaching+ & JoCoR & CREMA (ours)\\
\midrule
Symmetry-20\% &  $34.72 \pm 0.07$  & $52.11 \pm 0.21$  &  $50.48 \pm 0.24$ & $49.27 \pm 0.03$ & $53.41 \pm 0.09$ & $\textbf{57.21} \pm 0.25$\\
\midrule
Symmetry-50\% &  $16.86 \pm 0.09$  & $39.89 \pm 0.30$  &  $38.24 \pm 0.26$ & $40.04 \pm 0.70$ & $43.37 \pm 0.09$ & $\textbf{43.95} \pm 0.42$\\
\midrule
Symmetry-80\% &  $4.60 \pm 0.12$ & $16.08 \pm 0.15$  & $11.78 \pm 0.12$  & $13.44 \pm 0.37$ & $12.33 \pm 0.13$ & $\textbf{17.10} \pm 0.19$\\
\midrule
Asymmetric-40\% &  $26.93 \pm 0.10$  & $32.81 \pm 0.23$  & $33.36 \pm 0.28$  & $33.62 \pm 0.39$ & $32.66 \pm 0.13$ & $\textbf{38.61} \pm 0.25$\\
\midrule
Pairflip-40\% &  $27.48 \pm 0.12$  & $33.83 \pm 0.52$  & $33.94 \pm 0.18$  &  $33.80 \pm 0.25$ & $33.89 \pm 0.12$ & $\textbf{38.06} \pm 0.34$\\
\midrule
Pairflip-45\% &  $24.21 \pm 0.11$ &  $29.01 \pm 0.28$  & $29.57 \pm 0.15$  &  $26.93 \pm 0.34$ & $28.83 \pm 0.10$  & $\textbf{32.50} \pm 0.29$\\
\bottomrule
\end{tabular}
}
\caption{\small
Average test accuracy (\%) on \textsl{CIFAR-100} over the last ten epochs.}
\vspace{-2em}
\label{tab:cifar100}
\end{table*}

\begin{table}
\centering
\begin{minipage}[b]{0.48\linewidth}
    	
    \centering
	\scriptsize
	\begin{tabular}	{l | ccc | c }
		\toprule	 	
		    \multirow{2}{*}{Method} & \multicolumn{3}{c|}{Category} & \multirow{2}{*}{Accuracy} \\\cline{2-4}
		     & LA & LC & ND \\ \hline 
			\midrule			
			Cross-Entropy & & & & 79.4 \\
			ActiveBias~\cite{active-bias} & \checkmark & & & 80.5 \\
			PLC~\cite{PLC}   & & \checkmark & & 83.4 \\
			Co-teaching~\cite{Co-teaching}   & & & \checkmark & 80.2 \\
			SELFIE~\cite{SELFIE}  & & \checkmark & \checkmark & 81.8\\
			\midrule
			CREMA~(Ours) & \checkmark &  \checkmark & \checkmark & \textbf{84.2}\\
			
	    \bottomrule
	\end{tabular}
	\caption
		{
		\small	
		Test accuracy on Animal-10N. ``LA", ``LC" and ``ND" denote ``Loss Adjustment", ``Label Correction" and ``Noisy sample Detection" respectively.
		}
	\label{table:animal}
	
    \centering
    \scriptsize
	\begin{tabular}{l|cc}
		\toprule
		\multirow{2}{*}{Method}  & \multicolumn{2}{c}{Noise rates $\tau $} \\ \cmidrule{2-3} 
		 & Sym-20\% & Sym-40\% \\ \midrule
		Standard         & $74.08 \pm 0.23$    & $58.37 \pm 0.26$ \\ 
		PENCIL~\cite{PENCIL}	  & $73.73 \pm 0.21$   & $58.07 \pm 0.30$ \\ 
		Co-teaching~\cite{Co-teaching}       & $82.07 \pm 0.07$    & $73.25 \pm 0.19$ \\
		Co-teaching+~\cite{Co-teaching-PP}        & $82.27 \pm 0.23$    & $53.56 \pm 3.04$ \\
		JoCoR~\cite{JOCOR}   & $84.82 \pm 0.07$    & $76.12 \pm 0.17$ \\ 
		CREMA (ours)     & $\textbf{86.44} \pm 0.04$    & $\textbf{78.39} \pm 0.14$   \\
		\bottomrule
	\end{tabular}
		\caption{\small	
		\mm{Average test accuracy (\%) on IMDB dataset over the last ten epochs.}}
		\vspace{-3em}

	\label{tab:IMDB}

\end{minipage}
\hfill
\begin{minipage}[b]{0.48\linewidth}
    \scriptsize
    \centering
	\begin{tabular}	{l | ccc | c }
		\toprule	 	
		    \multirow{2}{*}{Method} & \multicolumn{3}{c|}{Category} & \multirow{2}{*}{Accuracy} \\\cline{2-4}
		     & LA & LC & ND \\ \hline 
			\midrule			
			Cross-Entropy & & & & 69.21 \\
			GCE~\cite{GCE} & \checkmark & & & 69.75 \\
			IMAE~\cite{IMAE} & \checkmark & & & 73.20 \\
			SCE~\cite{SCE} & \checkmark & & & 71.02 \\
			DM~\cite{DM} & \checkmark & & & 73.30 \\
			F-correction~\cite{forward-correction}  & \checkmark & & &69.84\\	
			M-correction~\cite{Arazo-BMM}  & \checkmark & & &  71.00 \\
			Masking~\cite{masking} & \checkmark & & &  71.10 \\
			Joint-Optim~\cite{tanaka2018joint}    & & \checkmark & & 72.23\\	
			Meta-Cleaner~\cite{MetaCleaner}  & & \checkmark &  & 72.50\\
			Meta-Learning~\cite{meta-learning-cvpr19}   & & \checkmark&  & 73.47\\
			PENCIL~\cite{PENCIL}  & & \checkmark & &73.49\\
			PLC~\cite{PLC}   & & \checkmark & & 74.02 \\
			Self-Learning~\cite{iccv19-deep-self}   & &  \checkmark & & 74.45 \\
			ProSelfLC~\cite{wang2021proselflc}  & &  \checkmark & & 73.40 \\
			Co-teaching~\cite{Co-teaching}   & & & \checkmark & 70.15 \\
			JoCoR~\cite{JOCOR}   & & & \checkmark & 70.30 \\
			C2D~\cite{C2D}   & & & \checkmark & 74.30 \\
			DivideMix$^{\dagger}$~\cite{dividemix}  & & & \checkmark & 74.48\\
			\midrule
			CREMA~(Ours)   & \checkmark &  \checkmark & \checkmark &  \textbf{74.53}  \\
			
	    \bottomrule
	\end{tabular}
	\caption
		{
		\small	
		Comparison with state-of-the-art methods in test accuracy on Clothing1M. $\dagger$ means the result without model ensemble~\cite{li2021learning_iccv2021}.
		}
		\vspace{-3em}
       \label{table:clothing}
\end{minipage}
\end{table}

\subsection{Comparison with state-of-the-art methods}

\textbf{Results on synthetic noisy datasets.}
\mm{Table \ref{tab:mnist}, Table \ref{tab:cifar10}, Table \ref{tab:cifar100} and Table \ref{tab:IMDB}} show the detailed results of the proposed \textit{CREMA} and other methods in multiple synthetic noisy cases on \mm{four} widely used datasets, i.e., MNIST, CIFAR-10, CIFAR-100 \mm{IMDB}. Specifically, four state-of-the-art LNL methods \mm{that are highly related to our work} are chosen for comparison: PENCIL~\cite{PENCIL}, Co-teaching~\cite{Co-teaching}, Co-teaching+~\cite{Co-teaching-PP}, JoCoR~\cite{JOCOR}. Standard DNN training with cross entropy is also included as baseline. All the results of these methods are reproduced with their public code and suggested hyper-parameters for fair comparison. From these tables, we can observe that \mm{most of} these methods show better performance than Standard in the most natural Symmetry-20\% case, \mm{except for PENCIL on IMDB dataset,} which verifies their robustness. Among them, 
\mm{JoCoR performs much better over other compared methods.}
However, when it comes to Pairflip-40\% and Pairflip-45\% \mm{noisy cases on image datasets}, their performance drops significantly. On the contrary, the proposed \textit{CREMA} can achieve consistent improvements over other methods on \mm{four} benchmarks across various noise settings. In the Pairflip-40\% and Pairflip-45\% cases, the proposed method outperforms other baselines by a large margin. Specifically, \textit{CREMA} can achieve 16.44\% and 25.26\% improvement in accuracy over JoCoR on CIFAR-10. When dealing with extremely noisy scenario, e.g. Symmetry-80\%, \textit{CREMA} can also perform generally better than other compared methods.
The result demonstrates the superiority \mm{and generality} of the proposed robust learning method across various types and levels of label noise 
\mm{on multimedia (i.e., image and text) datasets}.

\textbf{Results on real-world noisy datasets.}
Experiments on real-world noisy datasets Clothing1M~\cite{Clothing1M}, Animal-10N~\cite{SELFIE} are also conducted to verify the effectiveness of the proposed method. The baseline methods are chosen from recently proposed LNL methods. Specifically, several loss adjustment methods, including GCE~\cite{GCE}, IMAE~\cite{IMAE}, SCE~\cite{SCE}, DM~\cite{DM}, F-correction~\cite{forward-correction}, M-correction~\cite{Arazo-BMM}, Masking~\cite{masking}, ActiveBias~\cite{active-bias}, label correction methods, including Joint-Optim~\cite{tanaka2018joint}, PENCIL~\cite{PENCIL}, Self-Learning~\cite{iccv19-deep-self}, PLC~\cite{PLC}, ProSelfLC~\cite{wang2021proselflc} noisy sample detection approaches, including Co-teaching \cite{Co-teaching}, JoCoR~\cite{JOCOR}, C2D~\cite{C2D}, DivideMix~\cite{dividemix} and a hybrid method SELFIE~\cite{SELFIE} are compared with the proposed method.
Table~\ref{table:clothing} and Table~\ref{table:animal} show results on two real-world noisy datasets respectively. On the large-scale Clothing1M dataset, \textit{CREMA} outperforms all compared methods.
Note that \textit{CREMA} follows the standard DNN training procedure, and is similar to other co-training methods~\cite{Co-teaching,Co-teaching-PP,JOCOR} in terms of training time since the time cost for sample credibility modeling is negligible compared with DNN update. 
\eccv{It is worth noting that the proposed method outperforms these co-teaching methods by a large margin. Indicating that the discarded samples by co-teaching methods are actually valuable, and \textit{CREMA} well utilized all training samples.}
The \mm{best test accuracy} is achieved by \textit{CREMA} among the compared methods in Animal-10N as well. The results indicate that the proposed method can work well on high noise level (i.e., Clothing1M) and fine-grained (i.e., Animal-10N) real-world noisy datasets.

\vspace{-1.5em}

\begin{table} 
\begin{minipage}[b]{0.43\linewidth}
    \centering
    
    \scriptsize
    \begin{tabular}{c|c}
    \toprule	
        Method  & Test Accuracy (\%)  \\
        \midrule
         Baseline &  72.81 \\
        + Selective label update & 73.25 \\
        + Sequential likelihood 
        & 74.00 \\
        + Stability measurement 
        & \textbf{74.53} \\ 
    \bottomrule
    
    \end{tabular}
    \caption{\small
    Ablation studies of each component within \textit{CREMA} on Clothing1M dataset.}
    \label{tab:ablation1}
\end{minipage}
\hfill
\begin{minipage}[b]{.43\linewidth} 

    \centering 
    
    \scriptsize
    \begin{tabular}{c|c}
    \toprule	
        Estimator  & Test Accuracy (\%)  \\
        \midrule
         BMM &  74.09 \\
         GMM &  \textbf{74.53} \\
    \bottomrule
   
    \end{tabular}
    \vspace{4px}
     \caption{ \small
     Investigations on different mixture models on Clothing1M dataset.}
    \label{tab:ablation_GMM}
\end{minipage} 
\end{table}
\vspace{-4em}



\begin{table}[]
\footnotesize
    \centering
   
    \begin{tabular}{c|c|c|c|c|c|c}
    \toprule	
        Length of sequence $n$  & 1 & 2& 3& 4& 5& 6 \\
        \midrule
         Test Accuracy (\%) &  73.25 & 73.99 & \textbf{74.53} & 74.40 & 74.27 & 73.96 \\
    \bottomrule
    \end{tabular}
    \vspace{4px}
     \caption{ \small
     Investigations on length of sequence $n$ on Clothing1M dataset.}
    \label{tab:ablation_lenPool}
\end{table}

\vspace{-4em}
\subsection{Ablation studies} \label{ablation}

$\bullet$ \textbf{Component Analysis.} 
\textit{CREMA} contains several important components, including selective label learning strategy, sequential likelihood $\log{P\left(\mathbf{L}^n_t|\mathbf{f}_{t-n}, y\right)}$ and stability measurement $C(\mathbf{L}^n_t, y)$. To verify the effectiveness of each component, we conduct experiments on large-scale noisy dataset Clothing1M. The baseline method is built upon a simple co-teaching framework~\cite{JOCOR} combined with global label correction schemes~(as in \cite{PENCIL}), without the credibility guided loss adjustment strategy. The results are shown in Table~\ref{tab:ablation1}, we can see that, conform to the observation in Fig.~\ref{fig:grad}, the proposed selective label learning strategy achieves better results compared with the global correction counterpart. The sequential likelihood and stability measurement further boost the model performance with 0.75\% and 0.53\% accuracy gain, this indicates that the proposed sequential sample credibility modeling can effectively combat hard noisy samples mixed with clean ones. With all the three key components above, \textit{CREMA} can achieve 74.53\% test accuracy on Clothing1M.

\begin{figure}[!t]
 \centering
\includegraphics[width=\textwidth]{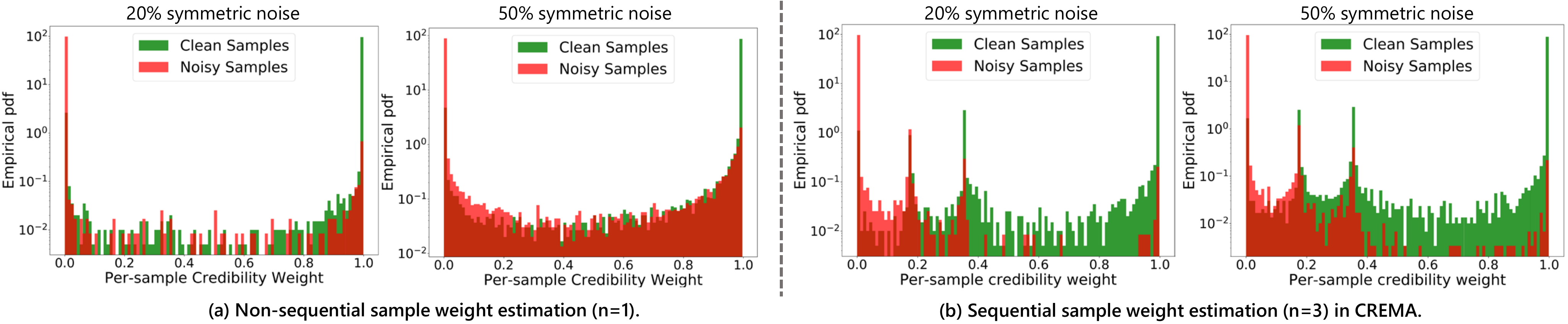}
  \vspace{-2em}
  \caption
  	{
  	\small
		Empirical PDF of the learned sample credibility weight between typical non-sequential (i.e., sequential length n=1) and sequential manner in CREMA. The model is trained on MNIST with 20\% and 50\% symmetric label noise. 
	  } 
\label{fig:gmm_weight}
\end{figure}

$\bullet$ \textbf{Length of sequence $n$.}
We also conduct experiments to investigate how the length of sequence $n$ affects the performance. Fig~\ref{fig:gmm_weight}(a) shows results on Clothing1M with various values~(in \{1, 2, 3, 4, 5, 6\}) of $n$. It can be observed that increasing the length of sequence helps achieve higher accuracy at first but turn poor after hitting the peak value. Intuitively, \lyxm{when} no temporal information is provided when $n=1$, \textit{CREMA} can not utilize consistency metric~\eccv{to identify hard noisy samples that blended with clean ones}, thus leading to an inferior result. When $n$ is larger than 4, we also notice that performance degrades, this is probably due to unreliable model inside the very long sequence can harm sample credibility modeling and reversely hinder the final result.     

$\bullet$ \textbf{Effect of different estimators.}
The probabilistic 
model plays the role of estimating the conditional probability $P\left(\mathbf{f}_{t}|\mathbf{f}_{t-1}, y\right)$ in Eq~(\ref{eq:derive}). We compare two different estimators, Gaussian Mixture Model (GMM)~\cite{GMM} and Beta Mixture Model (BMM)~\cite{Arazo-BMM} on Clothing1M. Table~\ref{tab:ablation_GMM} shows the results. We can see that GMM obtains a relatively higher test accuracy, but BMM can also achieve good results (74.09\%) as well. This indicates that the choice of normalized mixture model is not sensitive to the final result.

$\bullet$ \textbf{Reliability of the estimated sample credibility.}
Sample credibility plays the role of dynamical weight to modulate the loss term of each training sample, as in Eq~(\ref{eq:loss_reweighting}). In Fig.~\ref{fig:gmm_weight} we visualize the empirical PDF of the learned credibility weight of all training samples \eccv{between (b) the sequential estimation manner within \textit{CREMA} and (a) its non-sequential counterpart (i.e., $n=1$)} \eccv{under two different noisy settings}. It can be observed that the overall credibility weight of training samples is distinguishable for clean and noisy data in (b). Specifically, clean samples possess larger weight, thus contributing more \eccv{gradients during the process of} DNN training. Noisy samples are assigned with a relatively small weight to alleviate their negative impact. \eccv{However, the non-sequential weights yield significantly more samples that cannot be correctly-separated via a fixed threshold. Indicating that the proposed fine-grained sequential credibility estimation is more effective for reliable sample weight modeling. }


\section{Conclusion} \label{Conclusion}
In this paper, we propose a novel end-to-end robust learning method, called \textit{CREMA}. Towards the problem that previous works lack the consideration of intrinsic difference among difficulties of noisy samples. We follow the idea of divide-and-conquer that \lyxt{separates} clean and noisy samples via estimating the credibility of each training sample. Two branches are designed to handle the imperfectly separated sample \lyxt{sets} respectively. For easily recognizable noisy samples, we apply a selective label correction scheme avoiding erroneous label updates on clean samples. For hard noisy samples blended with clean ones, likelihood estimation of historical credibility sequence adaptively \lyxt{modulates} the loss term of each sample during training. Extensive experiments conducted on several synthetic and real-world noisy datasets verify the superiority of the proposed method.

\bibliographystyle{splncs04}
\bibliography{egbib}
\end{document}